\documentclass[conference,onecolumn]{IEEEtran}
\IEEEoverridecommandlockouts
\usepackage{cite}
\usepackage{amsmath,amssymb,amsfonts}
\usepackage{algorithmic}
\usepackage{graphicx}
\usepackage{textcomp}
\usepackage{acro}
\usepackage{todonotes}

\usepackage[utf8]{inputenc}
\usepackage[T1]{fontenc}
\usepackage[upright]{fourier}
\usepackage{tkz-kiviat,numprint}
\usepackage{capt-of}
\usepackage{pgfplots}
\usetikzlibrary{arrows}
\usetikzlibrary{shapes}

\newenvironment{customlegend}[1][]{%
	\begingroup
	\csname pgfplots@init@cleared@structures\endcsname
	\pgfplotsset{#1}%
}{%
	\csname pgfplots@createlegend\endcsname
	\endgroup
}%

\def\addlegendimage{\csname pgfplots@addlegendimage\endcsname}

\usetikzlibrary{positioning}

\DeclareAcronym{SDN}{short = SDN , long = software-defined networking}
\DeclareAcronym{NFV}{short = NFV , long = network function virtualization}
\DeclareAcronym{AGV}{short = AGV , long = automated guided vehicle}
\DeclareAcronym{UAV}{short = UAV , long = unmanned aerial vehicle}
\DeclareAcronym{UV}{short = UV , long = unmanned vehicle}
\DeclareAcronym{AP}{short = AP , long = access point}
\DeclareAcronym{MEC}{short = MEC , long = mobile edge computing}
\DeclareAcronym{LTE}{short = LTE , long = long term evolution}
\DeclareAcronym{LTE-A}{short = LTE-A , long = long term evolution-Advanced}
\DeclareAcronym{3GPP}{short = 3GPP , long = third generation partnership project}
\DeclareAcronym{5G-PPP}{short = 5G-PPP , long =  5G Public-Private Partnership}
\DeclareAcronym{VDE}{short = VDE , long = Verband Deutscher Elektrotechniker }
\DeclareAcronym{MTC}{short = MTC , long = machine-type communications}
\DeclareAcronym{eLTE}{short = eLTE , long = enterprise LTE}
\DeclareAcronym{mMTC}{short = mMTC , long = massive MTC}
\DeclareAcronym{URLLC}{short = URLLC , long = ultra reliable low latency communication}
\DeclareAcronym{SLAM}{short = SLAM , long = simulataneous localization and mapping}
\DeclareAcronym{NR}{short = NR , long = new radio}
\DeclareAcronym{LCP}{short = LCP , long = logical channel prioritization}
\DeclareAcronym{TS}{short = TS , long = technical specification}
\DeclareAcronym{NaaS}{short = NaaS , long = network as a service}
\DeclareAcronym{OF}{short = OF , long = OpenFlow}
\DeclareAcronym{IP}{short = IP , long = internet protocol}
\DeclareAcronym{MAC}{short = MAC , long = medium access control}
\DeclareAcronym{PDCP}{short = PDCP , long = packet data convergence protocol}
\DeclareAcronym{RFID}{short = RFID , long = radio frequency identification}
\DeclareAcronym{WLAN}{short = WLAN , long = wireless local area network}
\DeclareAcronym{IWLAN}{short = IWLAN , long = industrial WLAN}
\DeclareAcronym{PCF}{short = PCF, long = point coordination function}
\DeclareAcronym{DCF}{short = DCF, long = distributed coordination function}
\DeclareAcronym{HCF}{short = HCF, long = hybrid coordination function}
\DeclareAcronym{EDCA}{short = EDCA, long = enhanced distributed channel access}
\DeclareAcronym{HCCA}{short = HCCA, long = HCF controlled channel access}
\DeclareAcronym{REF}{short = REF , long = range extension function}
\DeclareAcronym{PRP}{short = PRP , long = parallel redundancy protocol}
\DeclareAcronym{GPS}{short = GPS , long = global positioning system}
\DeclareAcronym{GLONASS}{short = GLONASS , long = global navigation satellite system}
\DeclareAcronym{INS}{short = INS , long = global inertial navigation system}
\DeclareAcronym{GNSS}{short = GNSS , long = global navigation satellite system}
\DeclareAcronym{QoS}{short = QoS , long = quality of service}
\DeclareAcronym{RRC}{short = RRC , long = radio resource control}
\DeclareAcronym{PDU}{short = PDU , long = protocol data unit}
\DeclareAcronym{UE}{short = UE , long = user equipment}
\DeclareAcronym{TTI}{short = TTI , long = transmission time interval}

\begin{document}
	
	\title{Enabling Communication Technologies for Automated Unmanned Vehicles in Industry 4.0 \\
		\thanks{\textsuperscript{$\dagger$} Hans D. Schotten is also with the Intelligent Networks Research Group, German Research Center for Artificial Intelligence (DFKI), Kaiserslautern, Germany. (Hans$\_$Dieter.Schotten@dfki.de) }
	}
	
	\author{
		\IEEEauthorblockN{Amina Fellan, Christian Schellenberger, Marc Zimmermann, and Hans D. Schotten\textsuperscript{$\dagger$}}
		\IEEEauthorblockA{\textit{Institute for Wireless Communication and Navigation} \\
			\textit{Technische Universit{\"a}t Kaiserslautern}\\
			Kaiserslautern, Germany \\
			\{fellan, schellenberger, zimmermann, schotten\}@eit.uni-kl.de
		}
	}
	
	\maketitle
	
	\begin{abstract}
		Within the context of Industry 4.0, mobile robot systems such as \acp{AGV} and \acp{UAV} are one of the major areas challenging current communication and localization technologies. Due to stringent requirements on latency and reliability, several of the existing solutions are not capable of meeting the performance required by industrial automation applications. Additionally, the disparity in types and applications of \ac{UV} calls for more flexible communication technologies in order to address their specific requirements. In this paper, we propose several use cases for \acp{UV} within the context of Industry 4.0 and consider their respective requirements. We also identify wireless technologies that support the deployment of \acp{UV} as envisioned in Industry 4.0 scenarios.   
		
	\end{abstract}
	
	
	\section{Introduction}
	
	Under the umbrella of the Industry 4.0 paradigm, wireless communication is gaining more importance in industrial environments. It drives several of the planned solutions for intelligent manufacturing scenarios in smart factories. One of the suggested scenarios are material handling systems which employ \acp{AGV} for transportation of tools and products within a production facility from one location to another \cite{SD13}. Another possibility to transport smaller goods is to employ UAVs for this task \cite{uav_logistics}. Such applications place very high requirements on latency and reliability of the communication links which many of the existing wireless technologies struggle to satisfy. Also, the proposed wireless technologies have to be flexible enough to account for the wide range of \acp{UV} used in the industry. For instance, \acp{UAV}, also commonly known as drones, in logistics applications have completely different requirements in comparison with ground-based \acp{AGV}. While both types of \acp{UV} require highly reliable communication and accurate localization, \acp{UAV} control is more challenging due to their higher degrees of freedom and higher speed. Therefore they require lower latency for the communication technology used \cite{uav_latency}. \acp{AGV} on the other hand, might require higher data rates for video transmission in remote control applications \cite{3gpp_22804}.{\let\thefootnote\relax\footnote{{This is a preprint, the full paper has been accepted by the International Conference on Information and Communication Technology Convergence (ICTC), Jeju Island, Korea, October 2018}}}
	
	For an efficient operation of \acp{UV}, path planning is essential. Ideally, path planning should optimize the \acp{UV} routes by minimizing the distance traveled, minimizing the travel time or maximizing the utilization of \acp{UV} \cite{AK06}. Depending on how they navigate through a facility, for instance, \Acp{AGV} can operate in either of the following modes: fixed-guide-paths, or open-path. The former requires embedding physical guidance paths, usually using inductive or optical guides, in the factory floors. Whereas the open-path approach allows \acp{AGV} to roam freely in their environment, therefore adjusting to changes in their surroundings \cite{AK06}. 
	
	\par Earlier \acp{AGV} in industrial applications used a fixed-path induction-based guidance system, where wires are buried in the factory floor and a floor controller sends an electric current through the wires. An \ac{AGV} can detect and follow the currents in the wired paths \cite{GT87}. Other fixed-guide-path systems include magnetic or optic tapes to mark the ground paths. Without additional sensors, these systems allow the \acp{AGV} to follow a predetermined path. However, the \acp{AGV} do not have knowledge of where they are located. This can be rectified by having the location information encoded in \ac{RFID} tags next to the tracks. With the help of techniques such as odometry it is possible for the \ac{AGV} to acquire refined position estimates \cite{CDL10}. These systems while offering robustness, they suffer from inflexibility and relatively high-cost of their infrastructure.
	
	\par In contrast to the fixed-guide-path systems, open-path systems allow vehicles to move freely inside a facility. The prerequisites for such systems are a method to localize the \ac{AGV} and a map. There are different systems for a vehicle to localize itself indoors. One possibility is \ac{RFID} tags with position information distributed over the factory floor. The location can be estimated using gyroscopic sensors for angular information and wheel encoders for the distance \cite{CLL08}. Other existing positioning systems are based on laser sensors and reflective beacons with known positions. \acp{AGV} equipped with laser sensors or cameras can also use landmark identification schemes to determine their position \cite{DRN96}. All these systems, however, need a map of their environment in order to calculate the path they need to take. These systems are more flexible than fixed guide-path schemes, but require more expensive hardware, more involved coordination, and more computational resources. Wireless communication between the \acp{UV} themselves (UV-UV) and the production facility's network (UV-infrastructure) can improve global path planning and facilitate coordination amongst the \acp{UV}. Thus, it is essential for a smooth operation of future production facilities.
	
	In Section II, we first introduce existing use cases for \acp{AGV} and future ones anticipated in Industry 4.0 scenarios. The new use cases bring along a new set of requirements. We identify them in light of specifications set by the \ac{3GPP}. We provide an overview of current wireless technologies and examples of their deployments in Section III. Additionally, we list the open challenges for existing technologies in Section IV, as well as introduce future solutions that are promising to fulfill the needs of \acp{UV} in smart factories in Section V.
	
	\section{Use Cases and Requirements}
	Future production facilities should be able to support on-demand product customization. Therefore, production lines will be supplemented by small mobile transport robots, upon which the unfinished products can be redirected for further modifications inside the factory. Thus, \acp{UV} are needed. \acp{UV} can transport single workpieces, whole production batches or can deliver the needed resources according to the production schedule \cite{SZ18}. The following use cases based on \cite{3gpp_22804}, \cite{VDE17Funk} and \cite{5GPPFotF}, reflect possible applications for \acp{UV} in industrial automation settings.

	\subsection{\textbf{Smart modular production}}
	Factories of the future are envisaged to be highly reconfigurable and connected such that they can easily adapt to customization of products whenever required. Smart factories shall support a modular paradigm \cite{smart_kl}, which renders them independent from the physical location. Short production lines can be disassembled, transported from one location to another, and reassembled in short time. Therefore, future production facilities will consist of compact discrete workstations that can perform single work steps. These modules can also be plugged together to build longer linear production processes. To support a smooth operation in smart factories, \acp{AGV} are essential for an undisrupted flow of components, products and workstations to their intended destinations within the facility.
	
	\subsection{\textbf{Factory automation}}
	Automation in factories continues to expand and evolve with the objective to realize smart and efficient production. Human workers' tasks under automation has shifted from performing the physical work task to making high-level decisions for the robotic systems to complete the work. In relation to material handling, workers will no longer have to drive vehicles, e.g., forklifts or cranes, loading from the high rack storages to the production machines. Instead, \acp{AGV} and \acp{UAV} are used. Future \acp{UV} will support smart algorithms for self-coordination and collaboration within a fleet to optimize their route planning and adjust there distribution in a production facility. For local coordination environments, such as a workshop, \acp{UV} can communicate amongst each other and with neighboring machines to minimize transportation time such that the closest \ac{UV} to a scheduled task will pick it up as requested by the machine. As for global coordination on the factory level, a higher-level fleet management system, that has the knowledge of the current \acp{UV} distribution and their respective tasks, can assign \acp{UV} to emerging tasks as needed while ensuring an optimal routing and movement on the factory floor.

	\subsection{\textbf{Material and inventory live-tracking}}
	Flexible manufacturing requires the ability to track materials', workpieces', and products' flow throughout the entire supply chain. Whereas it is possible to implement tracking on individual items while being transported from one location to another, usually via \ac{RFID} tags and reader gates \cite{rfid_gates}, it is more efficient to transmit tracking data of a batch of carried items through the \ac{UV} transporting it. Particularly when live-tracking is in question. 	
	
	\subsection{\textbf{Mapping and surveillance}} 
	In dynamic industrial environment, e.g., modular smart factory, \acp{AGV} and \acp{UAV} can be used to keep track of the workshop floor plan with the help of technologies like \ac{SLAM} \cite{uav_slam},\cite{agv_slam}, where the \ac{UV} constructs a map of its surroundings while it simultaneously estimates its current position using that map. Additionally, cameras and sensors mounted on an \ac{UV} can be employed for surveillance of workshops and premises to keep track of intruders to restricted areas on the factory ground. It is also possible to build a virtual representation of the factory. This “digital twin” can be used to detect when changes occur in the factory floor and predict possible issues that might cause a halt to the production line \cite{dig_twin}.\\
	
	\begin{table}
		\caption{Technical requirements for \acp{UV} in industry 4.0 by 3GPP \cite{3gpp_22804}}
		\begin{center}
			{\setlength{\extrarowheight}{5pt}
				\begin{tabular}{|c|c|}
					\hline
					\textbf{Requirement} & \textbf{Vertical: Factories of the future} \\
					\hline
					Reliability (\%)	& 99,9999	  	 \\
					Latency (ms) 		& 40 - 500 *	 \\
					Speed (km/hr)		& up to 50		 \\   
					Data rate (Mbps) 	& up to 10 Mbps	 \\
					Coverage (m)		& 1000			 \\
					Number of \acp{UV}	& up to 100		 \\
					\hline 
					\multicolumn{2}{l}{$^{\mathrm{*}}$for standard UV operation.}
			\end{tabular}}
			\label{tab1}
		\end{center}
	\end{table}
	
	 \section*{Requirements}
\begin{figure}[h]
		
		\begin{tikzpicture}[scale=0.65, rotate=30]

		\begin{customlegend}[legend entries={Modular production,Factory automation,Inventory tracking, Mapping and Surveillance}, legend style={at={(-0.5,-8.5)},
			anchor=north east}]
		\addlegendimage{red,mark=square}
		\addlegendimage{blue,mark=*}
		\addlegendimage{green,mark=triangle}
		\addlegendimage{cyan,mark=o}
		\end{customlegend}

		\tkzKiviatDiagram[radial=6,lattice=5,gap=1,step=1,label space=1.5, font=\small]%
		{Reliability,
			Latency,
			Mobility,
			Data rate,
			Coverage,
			Number of UVs
		}
		\tkzKiviatLine[thick,color=red,mark=square,mark size= 5](4,2,3,2,2,2);
		\tkzKiviatLine[thick,color=blue,mark=*,mark size= 5](4,4,4,2,4,4);
		\tkzKiviatLine[thick,color=green,mark=triangle, mark size= 5](4,3,3,3,4,4);
		\tkzKiviatLine[thick,color=cyan,mark=o, mark size= 5](4,4,3,4,4,4);
		\tkzKiviatGrad[unity=1](4)

		\end{tikzpicture}
		\caption{Requirements for \acp{UV} in Industry 4.0}
		\label{fig:spiderweb}

	\end{figure}
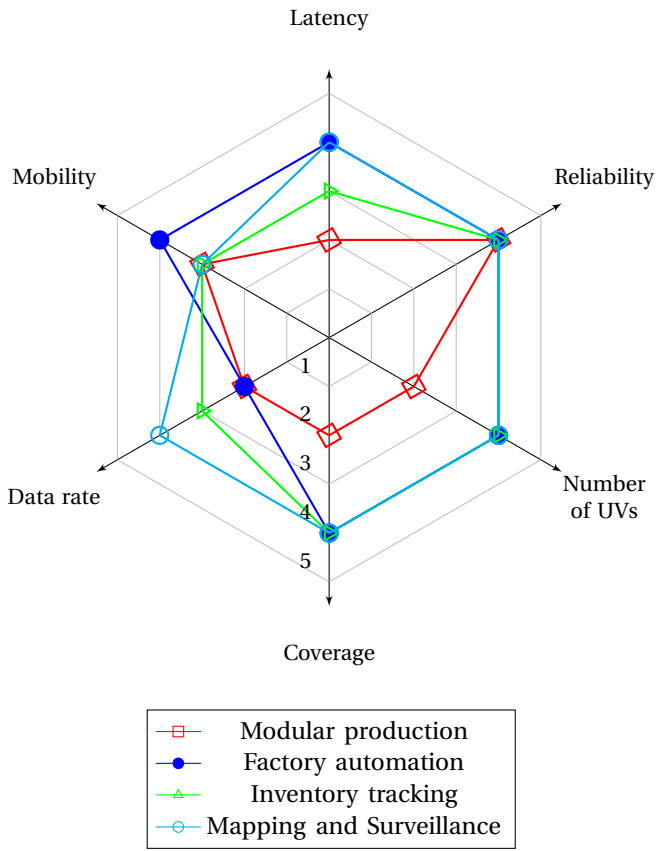

	In order to realize the suggested use cases for \acp{UV} in industrial environments, several requirements need to be accounted for when deciding on the enabling technologies. The \ac{3GPP} \ac{TS} 22.804 \cite{3gpp_22804} provides a study of automation in vertical domains for future 5G systems. We base the derivation of requirements for our suggested use cases on those set by the \ac{3GPP} for the "\textit{Factories of the Future}" use case in \cite{3gpp_22804} and listed in Table \ref{tab1}. The requirements for each use case are depicted in the radar chart shown in Figure \ref{fig:spiderweb}, where they are indicated and quantified on the axes with levels starting 0 for no requirement; 1 for a low requirement level; 2 for a medium requirement level; 3 for a high requirement level; and 4 for a very high requirement level. Reliability defines the maximum rate of packet loss tolerable for the application within the maximum tolerable end-to-end latency. For all suggested use cases involving \acp{UV} the requirement on reliability is set to higher than 99,9999 \%. Latency requirements vary for \acp{UV}; e.g., communications with machines as in the smart factory have very high requirements within 10 ms range, whereas for standard \acp{UV}' operation the requirement on latency is relaxed to up to 500 ms. All \acp{UV} use cases require mobility, however; the maximum speed possible while maintaining service reliability can vary from one use case to another. The higher the permissible speed, the tighter the execution of handover is within the required latency. The estimated data rate for the regular operation of \acp{UV} is within 1 Mpbs, except when video transmission is desired, as in surveillance, then higher data rates up to 1 Gbps need to be supported. The coverage area, where the communication services for \acp{UV} are supported, extends up to 500 m indoors and up to 1000 m for outdoor operation. The expected number of \acp{UV} is around 100 for a given use case. For the modular factory use case, the number of \acp{UV} required is limited compared to other use cases.

	
	\section{Existing Technologies}
	
	In this section, we discuss current wireless communication technologies that are used to support the operation of \acp{UV}.
	
	\subsection{\Ac{WLAN}}
	The IEEE 802.11 \ac{WLAN}, also known as WiFi, is considered to be one of the most ubiquitous wireless technologies in use. It was initially designed for operation in office buildings and home networks, providing high data rates (up to 433 Mbps for 802.11ac using single antenna \cite{cisco}) with no real-time guarantees. However, it was possible to employ it in various other areas that require real-time communications including industrial applications \cite{rt_wifi}, \cite{rt_wifi_2}, \cite{Siemens}. To ensure fulfillment of the rigid timing requirements, IEEE 802.11 is configured in such scenarios to operate in the \ac{PCF} mode; in which a point coordinator manages users' access to the medium. This centralized mode allows a deterministic operation for users. For instance, Siemens AG extended some of the features of the IEEE 802.11 standard to support the requirements of industrial communications \cite{Siemens}. By introducing \ac{IWLAN} with proprietary extensions to the 802.11 \ac{MAC} layer, which operates in the \ac{PCF} mode, and to protocols such as \ac{PRP}; it is possible to provide real-time communication with seamless roaming for terminals using PROFINET and EtherNet/IP industrial Ethernet standards. \ac{IWLAN} is designed for applications involving mobile robots, e.g. \acp{AGV}, cranes, and overhead conveyors. Researchers in \cite{wifi_limitation} compared the performance of \ac{PCF} to other IEEE 802.11 \ac{MAC} mechanisms in simulation. Namely, \ac{DCF}, \ac{EDCA}, and \ac{HCCA}, which is based on the \ac{HCF}. They considered conditions prevalent in industrial settings for real-time traffic. The results showed that the current IEEE 802.11 \ac{MAC} mechanisms struggle to support real-time traffic, particularly as the number of users and network traffic increases.

	\subsection{Cellular networks: \Ac{LTE} }
	As the requirement for the communication in the mobile users and inter-machine domains differ, the need to adjust the available cellular technologies arose, to better accommodate the tighter requirements, e.g., in terms of latency, for industrial automation. The 3GPP \ac{LTE} release 10 included the first provisions to enable \ac{MTC} over \ac{LTE} networks \cite{3gpp_rel10}. Within this context, \ac{MTC} technologies provide solutions to connect machines, sensors, and actuators to support their autonomous operation, as intended for the use case of \acp{UV}. \ac{LTE} can address several of the \ac{MTC} requirements thanks to its features. These include: low latency, high reliability, and wider coverage compared to other wireless technologies, e.g., WiFi \cite{lte_u}. 
	For instance, Huawei's adaptation of \ac{LTE}; \ac{eLTE}, which is based on \ac{LTE} in unlicensed bands and designed for industrial automation has been already deployed for the automated control of \acp{AGV} in Yangshan Port, China \cite{huawei}. 
	
	\subsection{\Acf{RFID} }
	\ac{RFID} technology is commonly used for inventory tracking and less frequently for localization and positioning applications. Researchers in \cite{rfid_cho} and \cite{rfid_lu}, studied and developed \ac{RFID}-based indoor location tracking systems for \acp{AGV}. For positioning applications, location information can be stored on the \ac{RFID} tags, which can be placed on the floors, ceilings, and throughout the area where \acp{UV} are operating. The \acp{UV} are equipped with an \ac{RFID} reader that detects the \ac{RFID} tags and uses the location information along with estimates of the distance between the tag and the reader to localize itself. The more tags an \ac{RFID} reader detects at a given location, the better the estimate. While localization using \ac{RFID} is an attractive solution for its relatively low cost and ease of deployment, it is susceptible to interference resulting from the surrounding environment.

	\section{Challenges}
	The Industry 4.0 paradigm raises plenty of new requirements on communication and localization technologies for \acp{UV} which existing solutions fall short to fulfill. Particularly within the fields of communication and localization technologies, demands are high on latency, data rate and handover mechanisms while simultaneously maintaining a good level of accuracy and reliability. We summarize in this section the open challenges, that \ac{UV} systems face, where new requirements for factory automation have been introduced. \\

	\noindent \textbf{Communication technologies}\\
	Low-latency and reliable communication technologies are indispensable for the operation of \acp{UV} in factory automation scenarios. Yet, current technologies lack a solution that can satisfy the wide range of requirements on data-rate, latency, reliability, scalability, and energy efficiency at once. Also, due to mobility of \acp{UV}, reliable and quick handovers are necessary such that data transmission is not disrupted when the \ac{UV} moves from the coverage of one cell/access-point to the next. When \acp{UAV} are in question, communication technologies must be energy efficient since the flying-time for a \ac{UAV} is strictly limited by its fuel supply or battery lifetime. Another great challenge that is prominent in industrial environments is  the hostility of propagation channels, due to electromagnetic interference caused by machinery and the common presence of reflectors in an industrial facility's infrastructure that enriches multipath components \cite{int_industry}. \\
	
	%
	
	\noindent \textbf{Localization systems}\\
	For outdoor localization scenarios of \acp{UV}, \ac{GNSS}-based localization is the norm. Current \ac{UAV} systems are using \ac{GPS}, \ac{GLONASS} receivers, or a combination of both, as well as \acp{INS} to estimate their position. However, indoor localization scenarios are much more challenging. \acp{GNSS} are not used since their receivers cannot acquire a signal indoors. Moreover, propagation conditions in industrial environments cause severe reflections and absorption of signals which produces inaccurate location estimates. As a result, path planning and \ac{SLAM} algorithms, that depend on the initial position estimate would be imprecise. Thus leading to difficulties for the \acp{UV} when navigating their vicinity, particularly when the environment is dynamically changing due to other participants in a workshop or factory floor. Therefore, a reliable localization technology has to be used in order to introduce AGVs in production facilities.\\
	
	\noindent \textbf{Other challenges}\\
	The heterogeneity of systems in a production facility creates the need for open-source based interfaces, protocols, and data formats that allow interoperability between the various systems for a functioning production line. This compatibility must be introduced starting from the top-level administrative and management tools to the bottom-level shop-floor machinery and device. In reality however, many of the current production systems have vendor-specific interfaces, protocols, and data formats which complicates the process of integrating and realizing flexible production. Additionally, future manufacturing systems have to be ideally backward-compatible with the current technologies. Since upgrades to the production systems' infrastructure are not only expensive but also time consuming. Thus, enabling communication and localization technologies for \acp{UV} should either make use of the existing wireless infrastructure, or be decoupled and designed such that modularity and ease of deployment are ensured.

	\section{Future 5G technologies}
	One of the major areas where future 5G technologies are envisioned to be applied is industrial networks. Service requirements, coupled with the operational conditions,  in industrial networks constitute a challenging setting for current wireless technologies. In the use case of \acp{UV} in particular, the requirements on reliable communication and relatively low latency are high.  
	
	\subsection{5G \ac{NR}}
	The next generation of communication technologies, 5G, is still under ongoing definition. 5G technologies are designed while keeping flexibility and scalability in mind to account for the requirements of future use cases. 5G \ac{NR} is the newer generation of radio interface and protocols for 5G networks proposed by the \ac{3GPP}.  In the latest \ac{3GPP} \ac{TS} \cite{3gpp_rel15}, \ac{URLLC} is supported as one of the use cases of verticals for 5G \ac{NR}. \acp{UV} are considered within \ac{URLLC} use cases. The special adjustments in 5G \ac{NR} for supporting \ac{URLLC} include \ac{LCP} restrictions as well as enabling packet duplication at the \ac{LTE}'s \ac{MAC} and \ac{PDCP} layers, respectively. The \ac{RRC} is the entity responsible for configuring protocol layers in the network and \ac{UE}. For  instance, \ac{LCP} restrictions made by the \ac{RRC}, e.g., \ac{TTI} size, cell and numerology for logical channels \cite{3gpp_rel15}, aim to prioritize \ac{URLLC} data traffic. Packet duplication at the \ac{PDCP} is also carried out by the \ac{RRC} and it ensures reliability and minimizes latency by sending \ac{PDU} duplicates simultaneously over two distinct channels \cite{5g_nr}.

	\subsection{\ac{SDN}}
	\ac{SDN} is an emerging technology that simplifies management of networks by separating the control and data planes in traditional network devices, such as routers and switches \cite{onf}. This approach allows centralizing the forwarding decisions and network logic at the \ac{SDN} controller and confining the network devices to solely enforcing those decisions. The \ac{SDN} controller acquires a global view of the network which enables it, for instance, to detect changes in the network topology \cite{PA16}. Therefore, it can keep track of the constantly moving \acp{UV} through information contained in the \ac{OF} messages being exchanged, these include: the \ac{IP} and \ac{MAC} addresses, the switch, and the port number the \ac{UV} is associated with. This information is updated whenever a host, in our case \ac{UV}, joins or leaves the network. As a result, better path planning and resource management can be achieved with the help of \ac{SDN}.  
	
	\subsection{Network slicing and \ac{NFV}}
	A novel paradigm that is planned for future 5G networks offers \ac{NaaS} for businesses and industries. The network is divided logically into network slices, based on the services they support, while still using a shared physical infrastructure \cite{OL17}. These network slices are operated independently and further optimized for the specific requirements of the vertical users. This allows, for instance, the verticals to operate their own private networks. Giving them full control and the ability to customize the network according to their needs and \ac{QoS}. Network slicing depends heavily on virtualization of the network infrastructure components using \ac{NFV} concepts. It is also simplified via \ac{SDN} for control. \ac{NFV} abstracts network functions from underlying specialized networking hardware, allowing their deployment on general-purpose servers or on cloud computing infrastructures \cite{nfv}. Whereas \ac{SDN} technology facilitates management of the slices through its centralized control plane \cite{nfv}.

	\subsection{Edge computing}
	
	For tasks requiring powerful processing and large computation power, e.g., \ac{SLAM}, \acp{UV} might not have the sufficient resources to carry them out. Edge computing is needed for such applications. It is a recent approach that aims to bring the cloud computation power at the network edge. As a result, it supports offloading extensive computations and storing large amounts of sensors' or video data, thus saving a \ac{UV}'s energy as well as its computational resources and reducing the overall delay in the network for time-critical applications. For example, the authors in \cite{UAV_edge} developed and implemented a test-bed for crowd surveillance with facial recognition using \acp{UAV}. The performance of processing the captured videos on board of the \ac{UAV} versus offloading it to an edge computer showed the advantage of the latter approach.       
	
	%
	
	\section{Conclusions}
	In this paper we outlined enabling communication technologies for material handling using \acp{UV}, which is a crucial component for the advancement and realization of future production automation ecosystems. We considered various use cases for \acp{UV} within the context of the Industry 4.0 initiative and derived their respective requirements in accordance with those set by organizations such as \ac{VDE}, \ac{3GPP}, and \ac{5G-PPP}. The highest requirements for the \acp{UV}' use cases are on latency and reliability, which current wireless technologies, such as WiFi and \ac{LTE}, struggle to fulfill. We focused on the existing wireless communication technologies that support \ac{UV}s' operation and denoted the challenges facing \ac{UV} systems under industry 4.0. The design of future 5G technologies takes into consideration industrial scenarios for factories of the future. With the help of prospective 5G techniques such as 5G \ac{NR}, \ac{SDN}, and edge computing, vertical industries will have better control over their private network and could fine-tune their networks according to the specific needs of individual use cases and ensure the required level of \ac{QoS}.
	
	\section*{Aknowledgement}
	This work has been supported by the Federal Ministry of Education and Research (BMBF) (grant no. 16KIS0725K, 5GANG) and the Federal Ministry for Economic Affairs and Energy (BMWi) (grant no. 01MA17008, IC4F) of the Federal Republic of Germany. The authors alone are responsible for the content of this paper.

	\bibliography{literature}

\begin{thebibliography}{10}
\providecommand{\url}[1]{#1}
\csname url@samestyle\endcsname
\providecommand{\newblock}{\relax}
\providecommand{\bibinfo}[2]{#2}
\providecommand{\BIBentrySTDinterwordspacing}{\spaceskip=0pt\relax}
\providecommand{\BIBentryALTinterwordstretchfactor}{4}
\providecommand{\BIBentryALTinterwordspacing}{\spaceskip=\fontdimen2\font plus
\BIBentryALTinterwordstretchfactor\fontdimen3\font minus
  \fontdimen4\font\relax}
\providecommand{\BIBforeignlanguage}[2]{{%
\expandafter\ifx\csname l@#1\endcsname\relax
\typeout{** WARNING: IEEEtran.bst: No hyphenation pattern has been}%
\typeout{** loaded for the language `#1'. Using the pattern for}%
\typeout{** the default language instead.}%
\else
\language=\csname l@#1\endcsname
\fi
#2}}
\providecommand{\BIBdecl}{\relax}
\BIBdecl

\bibitem{SD13}
L.~Sabattini, V.~Digani, C.~Secchi, G.~Cotena, D.~Ronzoni, M.~Foppoli, and
  F.~Oleari, ``{Technological Roadmap to Boost the Introduction of AGVs in
  Industrial Applications},'' in \emph{{IEEE International Conference on
  Intelligent Computer Communication and Processing (ICCP)}}.\hskip 1em plus
  0.5em minus 0.4em\relax IEEE, 2013, pp. 203--208.

\bibitem{uav_logistics}
T.~R.~F. Cavalcante, I.~V. d.~Bessa, and L.~C. Cordeiro, ``{Planning and
  Evaluation of UAV Mission Planner for Intralogistics Problems},'' in
  \emph{2017 VII Brazilian Symposium on Computing Systems Engineering (SBESC)},
  Nov 2017, pp. 9--16.

\bibitem{uav_latency}
Y.~Zeng, R.~Zhang, and T.~J. Lim, ``Wireless communications with unmanned
  aerial vehicles: opportunities and challenges,'' \emph{IEEE Communications
  Magazine}, vol.~54, no.~5, pp. 36--42, May 2016.

\bibitem{3gpp_22804}
3GPP, ``{Study on Communication for Automation in Vertical Domains},'' {3rd
  Generation Partnership Project (3GPP)}, Technical Specification (TS) 22.804,
  12 2017, version 1.0.0.

\bibitem{AK06}
T.~Le-Anh and M.~De~Koster, ``A review of design and control of automated
  guided vehicle systems,'' \emph{European Journal of Operational Research},
  vol. 171, no.~1, pp. 1--23, 2006.

\bibitem{GT87}
R.~Gaskins and J.~M. Tanchoco, ``Flow path design for automated guided vehicle
  systems,'' \emph{International Journal of Production Research}, vol.~25,
  no.~5, pp. 667--676, 1987.

\bibitem{CDL10}
A.~Codas, M.~Devy, and C.~Lemaire, ``{Robot localization algorithm using
  odometry and RFID technology},'' \emph{IFAC Proceedings Volumes}, vol.~43,
  no.~16, pp. 569--574, 2010.

\bibitem{CLL08}
B.-S. Choi, J.-W. Lee, and J.-J. Lee, ``{Localization and map-building of
  mobile robot based on RFID sensor fusion system},'' in \emph{Industrial
  Informatics, 2008. INDIN 2008. 6th IEEE International Conference on}.\hskip
  1em plus 0.5em minus 0.4em\relax IEEE, 2008, pp. 412--417.

\bibitem{DRN96}
H.~Durrant-Whyte, D.~Rye, and E.~Nebot, ``Localization of autonomous guided
  vehicles,'' in \emph{Robotics Research}.\hskip 1em plus 0.5em minus
  0.4em\relax Springer, 1996, pp. 613--625.

\bibitem{SZ18}
C.~Schellenberger, M.~Zimmermann, and H.~D. Schotten, ``Wireless communication
  for modular production facilities,'' \emph{arXiv preprint arXiv:1804.08273},
  2018.

\bibitem{VDE17Funk}
``{Funktechnologien f{\"ur} Industrie 4.0},'' Verband der Elektrotechnik
  Elektronik Informationstechnik e.V., Position Paper, 2017.

\bibitem{5GPPFotF}
``{5G and the Factories of the Future},'' 5GPP, White Paper, 2015.

\bibitem{smart_kl}
D.~Gorecky and S.~Weyer, ``{SmartFactory KL System Architecture for Industrie
  4.0 Production Plants},'' {Technology Initiative SmartFactory KL e.V.}, white
  paper SF-1.1, 04 2016.

\bibitem{rfid_gates}
J.~Song, C.~Haas, C.~Caldas, E.~Ergen, and B.~Akinci, ``Automating the task of
  tracking the delivery and receipt of fabricated pipe spools in industrial
  projects,'' vol.~15, pp. 166--177, 03 2006.

\bibitem{uav_slam}
M.~Bryson and S.~Sukkarieh, ``{Co-operative Localisation and Mapping for
  Multiple UAVs in Unknown Environments},'' in \emph{2007 IEEE Aerospace
  Conference}, March 2007, pp. 1--12.

\bibitem{agv_slam}
S.~Dörr, P.~Barsch, M.~Gruhler, and F.~G. Lopez, ``{Cooperative longterm SLAM
  for navigating mobile robots in industrial applications},'' in \emph{2016
  IEEE International Conference on Multisensor Fusion and Integration for
  Intelligent Systems (MFI)}, Sept 2016, pp. 297--303.

\bibitem{dig_twin}
Q.~Qi and F.~Tao, ``Digital twin and big data towards smart manufacturing and
  industry 4.0: 360 degree comparison,'' \emph{IEEE Access}, vol.~6, pp.
  3585--3593, 2018.

\bibitem{cisco}
Cisco, ``{802.11ac: The Fifth Generation of Wi-Fi},'' 2018.

\bibitem{rt_wifi}
Y.~H. Wei, Q.~Leng, S.~Han, A.~K. Mok, W.~Zhang, and M.~Tomizuka, ``{RT-WiFi:
  Real-Time High-Speed Communication Protocol for Wireless Cyber-Physical
  Control Applications},'' in \emph{2013 IEEE 34th Real-Time Systems
  Symposium}, Dec 2013, pp. 140--149.

\bibitem{rt_wifi_2}
L.~Seno, G.~Cena, S.~Scanzio, A.~Valenzano, and C.~Zunino, ``{Enhancing
  Communication Determinism in Wi-Fi Networks for Soft Real-Time Industrial
  Applications},'' \emph{IEEE Transactions on Industrial Informatics}, vol.~13,
  no.~2, pp. 866--876, April 2017.

\bibitem{Siemens}
Siemens, ``{iFeatures - Special Industrial Functions for Wireless Applications
  },'' 09 2017.

\bibitem{wifi_limitation}
R.~Costa, P.~Portugal, F.~Vasques, C.~Montez, and R.~Moraes, ``{Limitations of
  the IEEE 802.11 DCF, PCF, EDCA and HCCA to handle real-time traffic},'' in
  \emph{2015 IEEE 13th International Conference on Industrial Informatics
  (INDIN)}, July 2015, pp. 931--936.

\bibitem{3gpp_rel10}
3GPP, ``{Service requirements for Machine-Type Communications (MTC)},'' {3rd
  Generation Partnership Project (3GPP)}, Technical Specification (TS) 22.368,
  06 2010, version 10.1.0.

\bibitem{lte_u}
R.~Zhang, M.~Wang, L.~X. Cai, Z.~Zheng, X.~Shen, and L.~Xie, ``{LTE-unlicensed:
  the future of spectrum aggregation for cellular networks},'' \emph{IEEE
  Wireless Communications}, vol.~22, no.~3, pp. 150--159, June 2015.

\bibitem{huawei}
\BIBentryALTinterwordspacing
Huawei, ``{eLTE Port AGV Communications},'' 2018. [Online]. Available:
  \url{http://www.huawei.com/minisite/iot/en/elte-port-agv-communications.html}
\BIBentrySTDinterwordspacing

\bibitem{rfid_cho}
J.~H. Cho and M.~W. Cho, ``{Effective Position Tracking Using B-Spline Surface
  Equation Based on Wireless Sensor Networks and Passive UHF-RFID},''
  \emph{IEEE Transactions on Instrumentation and Measurement}, vol.~62, no.~9,
  pp. 2456--2464, Sept 2013.

\bibitem{rfid_lu}
S.~Lu, C.~Xu, R.~Y. Zhong, and L.~Wang, ``{A RFID-enabled positioning system in
  automated guided vehicle for smart factories},'' \emph{Journal of
  Manufacturing Systems}, vol.~44, pp. 179--190, 2017.

\bibitem{int_industry}
P.~Stenumgaard, J.~Chilo, J.~Ferrer-Coll, and P.~Angskog, ``Challenges and
  conditions for wireless machine-to-machine communications in industrial
  environments,'' \emph{IEEE Communications Magazine}, vol.~51, no.~6, pp.
  187--192, June 2013.

\bibitem{3gpp_rel15}
3GPP, ``{NR; NR and NG-RAN Overall Description; Stage 2},'' {3rd Generation
  Partnership Project (3GPP)}, Technical Specification (TS) 38.300, 06 2018,
  version 15.2.0.

\bibitem{5g_nr}
J.~Rao and S.~Vrzic, ``{Packet Duplication for URLLC in 5G: Architectural
  Enhancements and Performance Analysis},'' \emph{IEEE Network}, vol.~32,
  no.~2, pp. 32--40, March 2018.

\bibitem{onf}
ONF, ``{Software-Defined Networking: The New Norm for Networks},'' {Open
  Networking Foundation (ONF)}, ONF white paper, 04 2012.

\bibitem{PA16}
F.~Pakzad, M.~Portmann, W.~L. Tan, and J.~Indulska, ``Efficient topology
  discovery in openflow-based software defined networks,'' \emph{Computer
  Communications}, vol.~77, pp. 52--61, 2016.

\bibitem{OL17}
J.~Ordonez-Lucena, P.~Ameigeiras, D.~Lopez, J.~J. Ramos-Munoz, J.~Lorca, and
  J.~Folgueira, ``{Network Slicing for 5G with SDN/NFV: Concepts,
  Architectures, and Challenges},'' \emph{IEEE Communications Magazine},
  vol.~55, no.~5, pp. 80--87, May 2017.

\bibitem{nfv}
B.~Han, V.~Gopalakrishnan, L.~Ji, and S.~Lee, ``Network function
  virtualization: Challenges and opportunities for innovations,'' \emph{IEEE
  Communications Magazine}, vol.~53, no.~2, pp. 90--97, Feb 2015.

\bibitem{UAV_edge}
N.~H. Motlagh, M.~Bagaa, and T.~Taleb, ``{UAV-Based IoT Platform: A Crowd
  Surveillance Use Case},'' \emph{IEEE Communications Magazine}, vol.~55,
  no.~2, pp. 128--134, February 2017.

\end{thebibliography}
	
\end{document}